\documentclass[10pt,twocolumn,letterpaper]{article}

\usepackage{wacv}
\usepackage{times}
\usepackage{epsfig}
\usepackage{graphicx}
\usepackage{amsmath}
\usepackage{amssymb}
\usepackage{caption}
\usepackage{multirow}
\usepackage{authblk}
\usepackage{float}
\wacvfinalcopy

\def\wacvPaperID{363} 

\ifwacvfinal
\def\assignedStartPage{1} 
\fi

\ifwacvfinal
\usepackage[breaklinks=true,bookmarks=false]{hyperref}
\else
\usepackage[pagebackref=true,breaklinks=true,colorlinks,bookmarks=false]{hyperref}
\fi

\ifwacvfinal
\else
\fi

\begin{document}

\title{One-shot Compositional Data Generation for \\Low Resource Handwritten Text Recognition}

\author[*,1]{Mohamed Ali Souibgui }
\author[*,1]{Ali Furkan Biten}
\author[*,1,2]{Sounak Dey}
\author[1]{Alicia Fornés}
\author[3,4]{Yousri Kessentini}
\author[1]{Lluis Gomez}
\author[1]{Dimosthenis Karatzas}
\author[1]{Josep Lladós}
\affil[1]{Computer Vision Center, UAB, Spain}
\affil[2]{Helsing AI, Germany}
\affil[3]{Digital Research Center of Sfax,  Tunisia}
\affil[4]{SM@RTS : Laboratory of Signals, systeMs, aRtificial Intelligence and neTworkS}
\affil[ ]{\tt\small  \{msouibgui, abiten, sdey, afornes, lgomez, dimos, josep\}@cvc.uab.cat}
\affil[ ]{\tt\small  sounak.dey@helsing.ai \space\space\space yousri.kessentini@crns.rnrt.tn}
\renewcommand\Authands{ and }


\maketitle

\begin{abstract}
Low resource Handwritten Text Recognition (HTR) is a hard problem due to the scarce annotated data and the very limited linguistic information (dictionaries and language models). For example, in the case of historical ciphered manuscripts, which are usually written with invented alphabets to hide the message contents. Thus, in this paper we address this problem through a data generation technique based on Bayesian Program Learning (BPL). Contrary to traditional generation approaches, which require a huge amount of annotated images, our method is able to generate human-like handwriting using only one sample of each symbol in the alphabet.
After generating symbols, we create synthetic lines to train state-of-the-art HTR architectures in a segmentation free fashion. Quantitative and qualitative analyses were carried out and confirm the effectiveness of the proposed method.
\end{abstract}

\section{Introduction}
Handwritten Text Recognition (HTR) systems are based on deep learning, and require a significant amount of annotated data to reach a satisfactory performance. However, such systems suffer in low resource scenarios. For example, data scarcity is a common problem when dealing with manuscripts with uncommon scripts or alphabets. 
\let\thefootnote\relax\footnotetext{* Equal contribution.}


Historical ciphered manuscripts \cite{megyesi2019decode} is a typical case of low resource handwritten text, where invented alphabets replace the known ones to encrypt the text and hide the content from undesired readers.
Nowadays, many handwritten ciphered documents exist in archives consisting of military reports, diplomatic letters, records of secret societies, etc. 
Recognizing and extracting the hidden information is of great interest from the point of view of cultural heritage and history. However, a manual transcription and cryptoanalysis is costly both in terms of time and human resources. Therefore, the whole process needs for automatic tools. 

Because of the absence of context information in terms of language models and dictionaries, the automatic decryption of historical ciphered manuscripts is separated in two stages: transcription (HTR) and decipherment. The transcription step, which is the goal of this study, is a hard task due to the scarce annotated data to train, the paper degradation (typical in historical documents), and the changing alphabet across different ciphered manuscripts.




A typical solution to the lack of data is to create more examples for training via data augmentation or synthetic data generation.
But these techniques~\cite{shorten2019} require training data. Moreover, and contrary to humans, deep learning models are known to fail on compositional nature of generation. Here by compositional we refer to the generation of more complex items from simpler components/primitives, a process that humans can successfully do from just a single example ~\cite{lake2015human}. Furthermore, generating data that covers the distribution of an alphabet using only few examples of each symbol is an hard task for deep learning models.  


In this work, we bridge between generation by compositionality and data scarcity by generating realistic samples to serve as ground truth to train HTR models. Concretely, our work is based on Bayesian Program Learning (BPL)~\cite{lake2015human}, which uses simple programs to create more complex structures compositionally (i.e. to build rich concepts from simpler primitives/programs). However, BPL \cite{lake2015human} was used to generate perfectly segmented symbols rather than sequences. This poses an important limitation for handwritten text recognition, because text is a sequence of joined characters, especially in cursive handwriting. To overcome this limitation, we propose to create realistic text lines from the BPL generated symbols. These symbols are generated starting with one single example of each symbol in the alphabet. As a result, the generated handwritten text lines can be used to train data-hungry HTR deep learning models for manuscripts with rare alphabets.
As a study case, we use the Borg \footnote{\url{https://cl.lingfil.uu.se/~bea/borg/}} cipher, a historical ciphered manuscript, written with an invented alphabet and containing many touching symbols, as shown in Figure~\ref{fig:borg}. The goal is to transcribe such a difficult text with the minimal user intervention (in terms of labelled data). 



As far as we know, this is the first work that effectively uses BPL for Handwritten Text Recognition, as an example of application of BPL for sequence recognition. The contributions of our work can be summarized as follows: 
\begin{itemize}
    \item We use BPL as a realistic symbol generation technique for handwriting recognition. The quantitative, qualitative results and human studies demonstrate its effectiveness. 
    \item We reduce the cost of annotation and human labor by automatically generating handwritten text lines by using just a single example per alphabet symbol.
    \item We experimentally show that the generated data benefits HTR models. Indeed, out approach outperforms the current state of the art on cipher recognition. This paves the way for it  to be applied for other low resource manuscripts.
     
\end{itemize}

The rest of the paper is organized as follows: related work is overviewed in Section \ref{section:relatedworks}. Then, BPL for handwriting generation is described in Section \ref{section:method}. Section~\ref{section:data_generation} describes how BPL can be successfully used to generate textlines with high but realistic handwriting style variability. Section~\ref{section:bpl_in_htr} analyzes the experimental results, whereas Section \ref{section:conclusion} presents the conclusions and future work.


\begin{figure*}[t!]
\centering
\begin{minipage}{.65\textwidth}
\centering
\includegraphics[width=.77\linewidth]{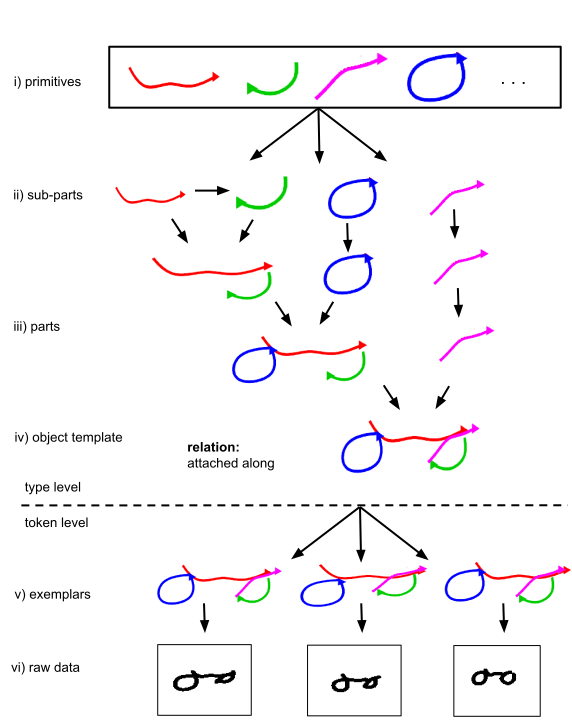}
\captionof{figure}*{\textbf{A}}

\end{minipage}%
\vline
\begin{minipage}{.2\textwidth}
\vspace{5mm}
\centering
\begin{tabular}{c}
 \fbox{\includegraphics[width=0.3\columnwidth]{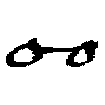}}\\
 \\
 \fbox{\includegraphics[width=0.3\columnwidth]{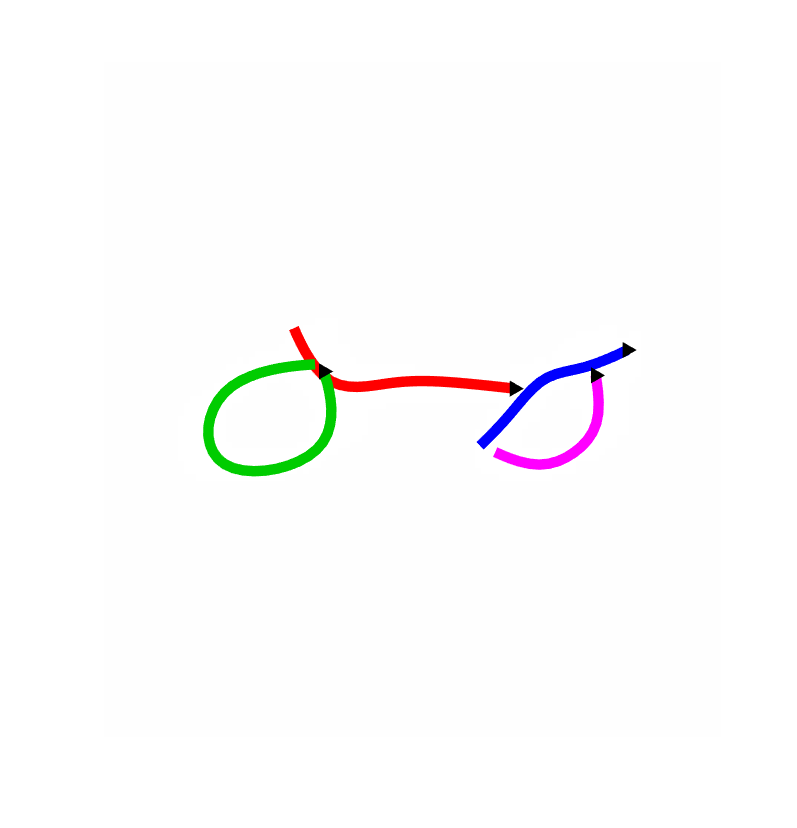}}\\
 -975.9\\
 \fbox{\includegraphics[width=0.3\columnwidth]{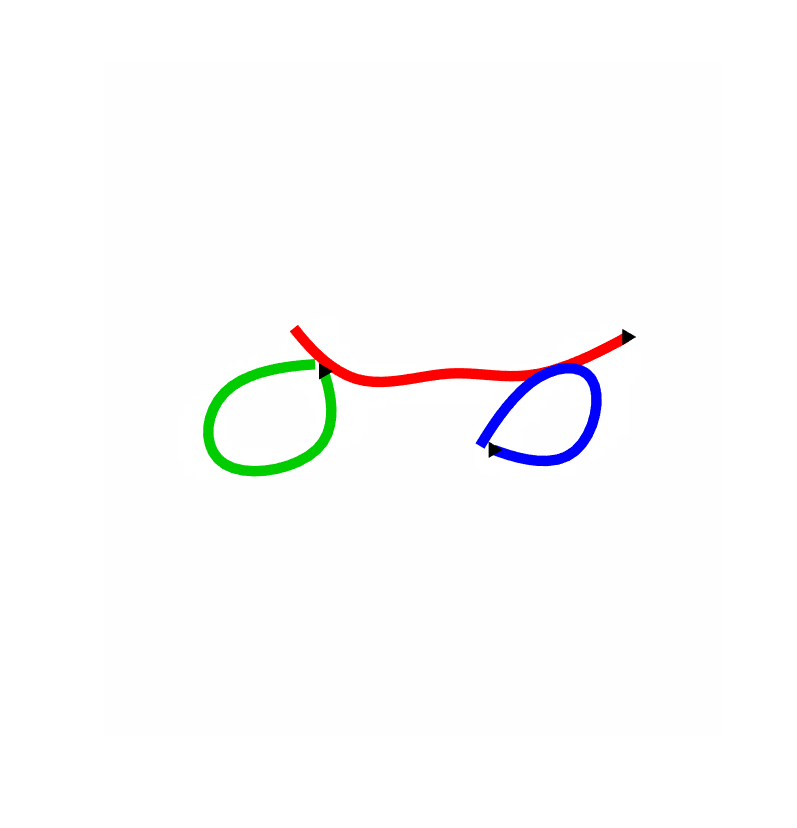}}\\
 -1024\\
 \fbox{\includegraphics[width=0.3\columnwidth]{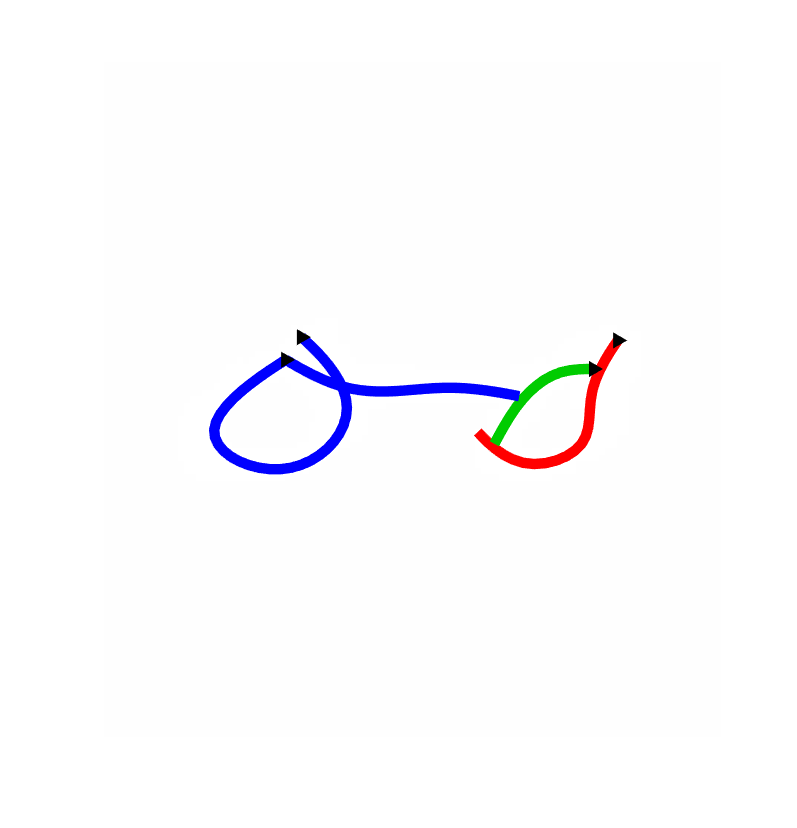}}\\
 -1091\\
 \fbox{\includegraphics[width=0.3\columnwidth]{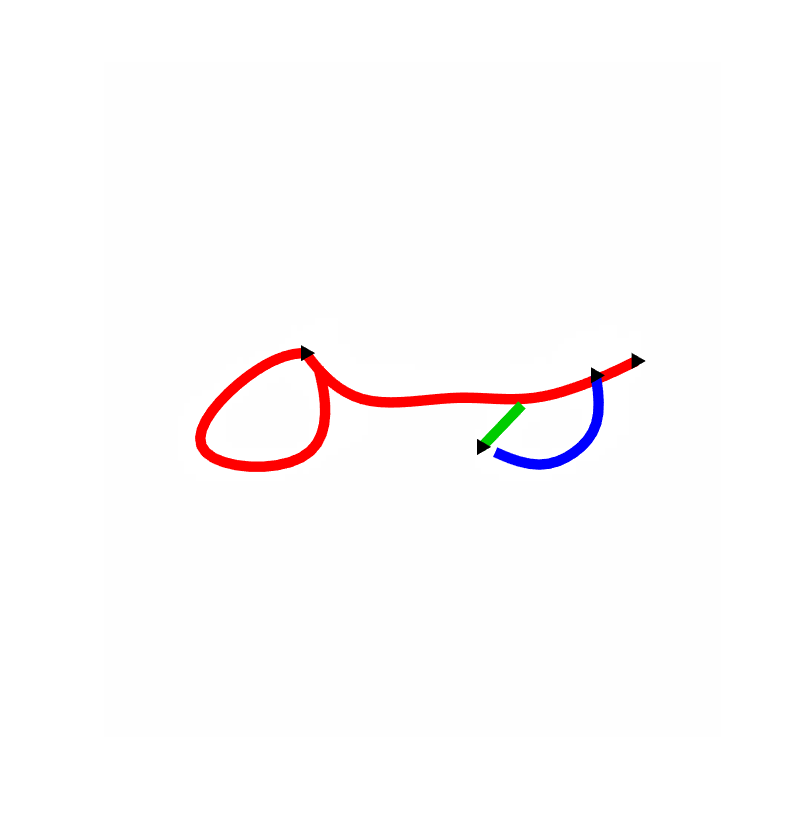}}\\
 -1204\\
 \fbox{\includegraphics[width=0.3\columnwidth]{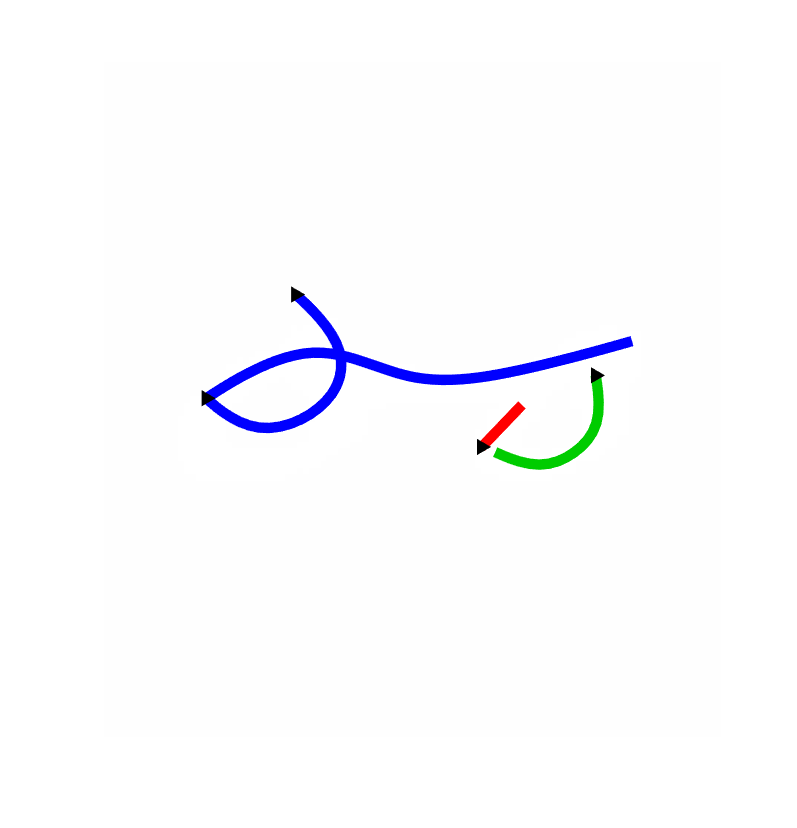}}\\
 -1754
\end{tabular}

\captionof{figure}*{\textbf{B}}
\label{fig:synth_bpl_parses}
\end{minipage}
\\
\caption{\textbf{(A)} A generative model of handwritten ciphered symbols. (i) From a library of color coded primitives, new types are generated, (ii) combining these subparts, (iii) to further generate parts, (iv) and then to define simple programs by combining parts with relations. (v) Running these programs new tokens are generated, (vi) which are then rendered as raw images.
\textbf{(B)} An image along with their log-probability scores for the five best programs. Parts are distinguished by color, with a colored flat back indicating the beginning of a stroke and a black arrowhead indicating the end. 
}
\label{fig:bpl_method}
\vspace{-0.4cm}
\end{figure*}

\section{Related Work}\label{section:relatedworks}

\subsection{Handwritten Ciphered Text Recognition}
Handwritten ciphered text recognition is related to the more general problem called HTR. The difference is that for ciphers, the alphabet is frequently unknown. In HTR, most of the existing approaches focus on well-known scripts \cite{jemni2020,kang2018convolve} using Convolutional Neural Networks (CNN) and Recurrent Neural Networks (RNN). These methods, despite their good results in HTR, cannot be used for ciphered text recognition because of the changing alphabets across ciphers, and the scarce annotated data for training. Indeed, an attempt to recognize ciphers based on RNNs \cite{fornes2017} showed that results were satisfying only when enough labeled data from the same cipher alphabet was available.
Some learning-free approaches have been proposed to cope with the lack of annotated data in ciphers \cite{baro2019,yin2019}. These methods are based on symbol segmentation and clustering, but their performance drops when symbols are difficult to segment (e.g. touching symbols). Lately, few-shot learning approaches for text recognition \cite{zhang2020,souibgui2020} have been introduced to deal with the few annotated data while keeping a good performance. They propose to transform the recognition problem into a matching problem, where the user is asked to provide only one or few examples of the alphabet, instead of annotating full textlines. Given those "shots", the system must output the transcribed text based on a similarity matrix between the provided alphabet and the text lines. 

\subsection{Data Augmentation and Generation for Handwritten Text}

Data augmentation and generation are suitable solutions for deep learning models when data is limited. Classic data augmentation techniques used some image manipulation tricks \cite{shorten2019}, such as geometric transformations (e.g. rotations, resizing, warping), random erasing, color transformations, font thickness, flipping, etc. However, these methods need training data, and the augmented text is moderately realistic.

In the case of online handwriting, trajectory reconstruction approaches were introduced based on the kinematic theory of human movements \cite{bhattacharya2017sigma,plamondon2006multi,plamondon1998generation} or by recurrent neural networks \cite{graves2013generating}.  However, online information (e.g. stroke trajectory, speed, pressure) is not available in historical manuscripts, where only text images are available. Generative adversarial networks \cite{goodfellow2014} and style transfer \cite{gomez2019selective,jing2019} methods were utilized to generate the handwritten text from images. In \cite{chang2018generating}, an approach to generate handwritten characters from an existing printed font was proposed. Also, in \cite{kang2020ganwriting}, cursive Latin words were generated conditioning on a content (text) and a writing style. 
But these approaches need a huge set of annotated data to be trained on for each particular handwriting style, which is not available for low resource applications. It is true that there are some attempts to use these techniques for few shot by using only few samples of each character class \cite{bartunov2018few,cha2020few,clouatre2019figr,srivatsan2019deep}. Nonetheless, the results are still moderate in terms of quality and most of the methods are focusing on font translation while keeping the same text shape as the example that are conditioned on.

In this work, and to overcome the above limitations, we explore the use of BPL~\cite{lake2015human} to mimic the human ability of generating new unseen characters, while maintaining high quality and shape variation, from a single example.

\section{Bayesian Program Learning (BPL)}\label{section:method}

Human beings have the ability to learn new concepts from a single example. Contrary, deep learning-based methods usually require tens or hundreds of examples to reach a human-level performance on recognition, generation, or parsing tasks. Thus, the generation of handwritten data from few examples is still challenging. The Bayesian Program Learning (BPL) introduced in~\cite{lake2015human} showed a great ability to learn rich concepts  compositionally and generate new examples from a \emph{single unseen} concept, making it an ideal solution for the data scarcity  problem.   

As it can be appreciated from Fig.~\ref{fig:bpl_method}-A, BPL works in a hierarchical manner. At the highest layer, there are two levels called type level and token level (the dashed line in the middle). Type level consist of 4 steps which are sampling primitives, sampling sub-parts, sampling sub-part sequence and sampling relation. In BPL, primitives are defined as the smallest stroke in unit and time. More specifically, given a held out set of data, BPL fits a Gaussian Mixture Model (GMM) on the normalized strokes according to their length and time information. Each of the center of GMM cluster is treated as primitives and used as a starting point. 

After obtaining the primitives, the number of primitives is sampled with $P(\kappa)$, where the distribution is obtained according to held out data. Then, the number of sub-parts with $P(n_i|\kappa)$ and sample sub-parts sequences with $P(S_i|S_1, S_2, ..., S(i_1))$ is sampled to decide which primitives should have relations, i.e. whether they should be combined into parts. Finally, BPL samples relations given sub-part sequence $P(R_i|S_1, S_2, ..., S(i_1))$ where there are four relations defined a priori for how the two strokes can be attached together. The two strokes can be attached ``along'', ``at start'', ``at end'', or ``independent''. All of the mentioned parts are combined with conditional probability to a program, $P(\psi)$

\begin{equation}
\label{e:bayes_deomposition}
P(\psi)=P(\kappa) \prod_{i=1}^{\kappa} P\left(S_{i}\right) P\left(R_{i} \mid S_{1}, \ldots, S_{i-1}\right)
\end{equation}

The token level parameters, referred as $\theta$, consist of global re-scaling and a global translation of the center of mass of each sub-part sequence. Moreover, BPL adds variance to the created types in terms of start location, trajectory, affine transformation so that the generated samples are as unique as possible. Token level parameters are distributed as $P(\theta|\psi)$ and the end product of token level is $P(I|\theta)$.

At the inference phase, it produces a new image $I^{(2)}$ given an image $I^{(1)}$. First, BPL reduces the line width of an image to one pixel and then runs a random walk algorithm to collect at most 10 parses of the $I^{(1)}$. An example of these parses can be found in Fig.~\ref{fig:bpl_method}-B. These parses are sorted according to log-probability of the random walk search and the most likely one is taken as the starting point. Afterwards, a new image is generated according to following formulation:

\begin{equation}\begin{aligned}
&P\left(I^{(2)}, \theta^{(2)} \mid I^{(1)}\right)=
\\
&\ \ \ \ \ \ \sum_{i=1}^{K} \sum_{j=1}^{N} \frac{w_{i}}{N} P\left(I^{(2)} \mid \theta^{(2)}\right) P\left(\theta^{(2)} \mid \psi^{[i j]}\right)
\end{aligned}\end{equation}

One of the main advantages of using BPL as a data augmentation is that it does not require huge training samples but more importantly, domain knowledge is minimized. For example, BPL can be trained on the Omniglot symbols, and later used on the Borg cipher symbols. Secondly, BPL can generate new exemplars from a \emph{single unseen} example, whereas deep models are incapable of for the moment. Finally and most importantly, the output images have enough variability while keeping the main structure to be used as a training set.
In all of our experiments, we have used the code\footnote{\url{https://github.com/brendenlake/BPL}} to generate each symbol.
The parsing includes a 'fast\_mode' option which skips the expensive procedure of fitting the strokes to the details of the image. 

\section{Handwritten Symbol Generation with BPL}\label{section:data_generation}
In this section, we present the generation of cipher symbols using BPL. We quantify the effectiveness of the method and include a discussion of qualitative and human study results.

\begin{figure}[t!]
    \centering
    \begin{tabular}{|l|}
      \hline
      \includegraphics[width=0.9\columnwidth, height=0.75cm]{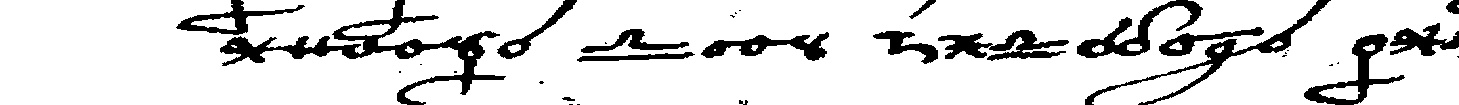}\\
      \hline
      \includegraphics[width=0.9\columnwidth, height=0.75cm]{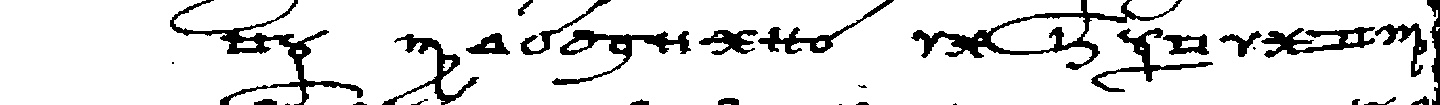}\\
      \hline
    \end{tabular}
    \caption{Two lines images from the Borg cipher. The image shows that there are frequent touching symbols in this manuscript, even between different lines.}
    \label{fig:borg}
    \vspace{-0.51cm}
\end{figure}
\subsection{Dataset}

Borg is a 408 pages ciphered manuscript belonging to the 17th century. Its alphabet is composed of abstract, esoteric symbols, Roman letters and some diacritics. Fig.~\ref{fig:borg} illustrates this handwritten text. As it can be seen, symbols are hard to segment, mainly because of the frequent symbol overlapping not only between consecutive symbols, but also between the different lines. Following related works \cite{baro2019,souibgui2020}, we have used 273 lines extracted from 16 pages for testing. Note that a pre-processing step (binarization and projections) has been applied to obtain those lines from the full pages manuscripts. For data generation, we manually cropped 10 samples of each class in the alphabet.

\begin{figure*}[t]
\begin{center}

\begin{tabular}{l@{\hskip 0.25in}lll}

\textbf{(A)}&
\begin{tabular}{lll}
\multicolumn{3}{c}{\includegraphics[width=0.08\columnwidth]{images/bpl/w/4.png}}                                                   \\ \hline
\multicolumn{1}{|l|}{\includegraphics[width=0.08\columnwidth]{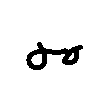}} & \multicolumn{1}{l|}{\includegraphics[width=0.08\columnwidth]{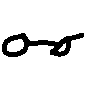}} & \multicolumn{1}{l|}{\includegraphics[width=0.08\columnwidth]{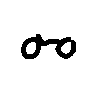}} \\ \hline
\multicolumn{1}{|l|}{\includegraphics[width=0.08\columnwidth]{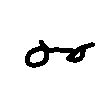}} & \multicolumn{1}{l|}{\includegraphics[width=0.08\columnwidth]{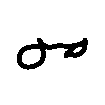}} & \multicolumn{1}{l|}{\includegraphics[width=0.08\columnwidth]{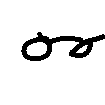}} \\ \hline
\multicolumn{1}{|l|}{\includegraphics[width=0.08\columnwidth]{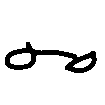}} & \multicolumn{1}{l|}{\includegraphics[width=0.08\columnwidth]{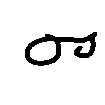}} & \multicolumn{1}{l|}{\includegraphics[width=0.08\columnwidth]{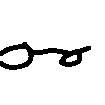}} \\ \hline
\end{tabular}
 &
\begin{tabular}{lll}
\multicolumn{3}{c}{\includegraphics[width=0.08\columnwidth]{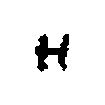}}                                                   \\ \hline
\multicolumn{1}{|l|}{\includegraphics[width=0.08\columnwidth]{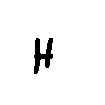}} & \multicolumn{1}{l|}{\includegraphics[width=0.08\columnwidth]{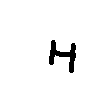}} & \multicolumn{1}{l|}{\includegraphics[width=0.08\columnwidth]{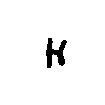}} \\ \hline
\multicolumn{1}{|l|}{\includegraphics[width=0.08\columnwidth]{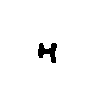}} & \multicolumn{1}{l|}{\includegraphics[width=0.08\columnwidth]{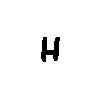}} & \multicolumn{1}{l|}{\includegraphics[width=0.08\columnwidth]{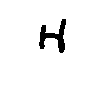}} \\ \hline
\multicolumn{1}{|l|}{\includegraphics[width=0.08\columnwidth]{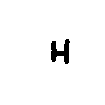}} & \multicolumn{1}{l|}{\includegraphics[width=0.08\columnwidth]{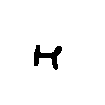}} & \multicolumn{1}{l|}{\includegraphics[width=0.08\columnwidth]{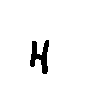}} \\ \hline
\end{tabular}

&

\begin{tabular}{lll}
\multicolumn{3}{c}{\includegraphics[width=0.08\columnwidth]{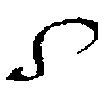}}                                                   \\ \hline
\multicolumn{1}{|l|}{\includegraphics[width=0.08\columnwidth]{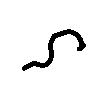}} & \multicolumn{1}{l|}{\includegraphics[width=0.08\columnwidth]{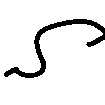}} & \multicolumn{1}{l|}{\includegraphics[width=0.08\columnwidth]{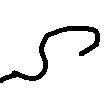}} \\ \hline
\multicolumn{1}{|l|}{\includegraphics[width=0.08\columnwidth]{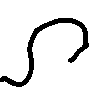}} & \multicolumn{1}{l|}{\includegraphics[width=0.08\columnwidth]{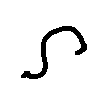}} & \multicolumn{1}{l|}{\includegraphics[width=0.08\columnwidth]{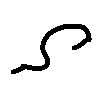}} \\ \hline
\multicolumn{1}{|l|}{\includegraphics[width=0.08\columnwidth]{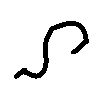}} & \multicolumn{1}{l|}{\includegraphics[width=0.08\columnwidth]{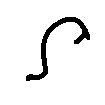}} & \multicolumn{1}{l|}{\includegraphics[width=0.08\columnwidth]{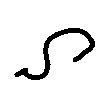}} \\ \hline
\end{tabular}

\cr
\noalign{\smallskip}\hline\noalign{\smallskip}
\textbf{(B)}&
\begin{tabular}{lll}
                       \hline
\multicolumn{1}{|l|}{\includegraphics[width=0.08\columnwidth]{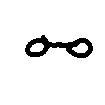}} & \multicolumn{1}{l|}{\includegraphics[width=0.08\columnwidth]{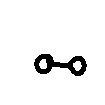}} & \multicolumn{1}{l|}{\includegraphics[width=0.08\columnwidth]{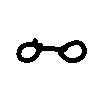}} \\ \hline
\multicolumn{1}{|l|}{\includegraphics[width=0.08\columnwidth]{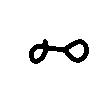}} & \multicolumn{1}{l|}{\includegraphics[width=0.08\columnwidth]{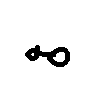}} & \multicolumn{1}{l|}{\includegraphics[width=0.08\columnwidth]{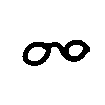}} \\ \hline
\multicolumn{1}{|l|}{\includegraphics[width=0.08\columnwidth]{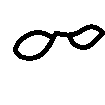}} & \multicolumn{1}{l|}{\includegraphics[width=0.08\columnwidth]{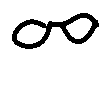}} & \multicolumn{1}{l|}{\includegraphics[width=0.08\columnwidth]{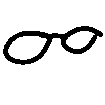}} \\ \hline
\end{tabular}
 &
\begin{tabular}{lll}
                       \hline
\multicolumn{1}{|l|}{\includegraphics[width=0.08\columnwidth]{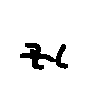}} & \multicolumn{1}{l|}{\includegraphics[width=0.08\columnwidth]{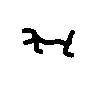}} & \multicolumn{1}{l|}{\includegraphics[width=0.08\columnwidth]{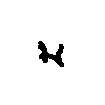}} \\ \hline
\multicolumn{1}{|l|}{\includegraphics[width=0.08\columnwidth]{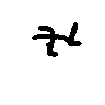}} & \multicolumn{1}{l|}{\includegraphics[width=0.08\columnwidth]{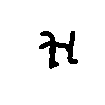}} & \multicolumn{1}{l|}{\includegraphics[width=0.08\columnwidth]{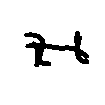}} \\ \hline
\multicolumn{1}{|l|}{\includegraphics[width=0.08\columnwidth]{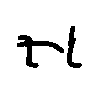}} & \multicolumn{1}{l|}{\includegraphics[width=0.08\columnwidth]{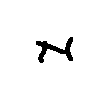}} & \multicolumn{1}{l|}{\includegraphics[width=0.08\columnwidth]{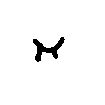}} \\ \hline
\end{tabular}

&

\begin{tabular}{lll}
                       \hline
\multicolumn{1}{|l|}{\includegraphics[width=0.08\columnwidth]{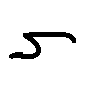}} & \multicolumn{1}{l|}{\includegraphics[width=0.08\columnwidth]{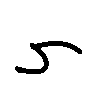}} & \multicolumn{1}{l|}{\includegraphics[width=0.08\columnwidth]{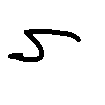}} \\ \hline
\multicolumn{1}{|l|}{\includegraphics[width=0.08\columnwidth]{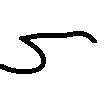}} & \multicolumn{1}{l|}{\includegraphics[width=0.08\columnwidth]{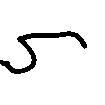}} & \multicolumn{1}{l|}{\includegraphics[width=0.08\columnwidth]{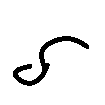}} \\ \hline
\multicolumn{1}{|l|}{\includegraphics[width=0.08\columnwidth]{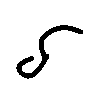}} & \multicolumn{1}{l|}{\includegraphics[width=0.08\columnwidth]{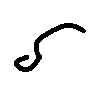}} & \multicolumn{1}{l|}{\includegraphics[width=0.08\columnwidth]{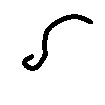}} \\ \hline
\end{tabular}

\end{tabular}
\end{center}
\caption{Generating new exemplars given one ciphered symbol. \textbf{(A):}  Conditioning on the same symbol (in-sample) shown on top of the nine-cipher grids. \textbf{(B):} Conditioning on a different example of the same class (out-sample). The nine-character grids were generated by BPL.}
\label{fig:synth_bpl}
\vspace{-0.5cm}
\end{figure*}

\subsection{Data generation results - Symbol Level}
In this section, we show the results of the BPL generation of Borg symbols and evaluate them according to a human study. 
\label{sec:qualitative}
\subsubsection{Qualitative Results}
We provide two types of qualitative results. In Fig.~\ref{fig:synth_bpl}-A, samples are generated using the top-right character. On the other hand, in Fig.~\ref{fig:synth_bpl}-B, any example belonging to the same class but the top character can be used for generation. In other words, the generated examples of Fig.~\ref{fig:synth_bpl}-B are not conditioned on the top character, rather it is conditioned on different examples from the same class of the top one.
We will call the former in-sample, and the latter out-of-sample examples.
 
As it can be seen for the in-sample generation from the Fig.~\ref{fig:synth_bpl}-A, BPL mostly keeps intact the symbol structure while introducing variety. For example, from the first column, it can be seen that some of the parts where it looks like ``o", BPL transforms it into a more open ``o'' (3rd row, 3rd column) and transforms the connection of lines. Moreover, it can change the line thickness and make the lines shorter or curved, see 3rd column of Fig.~\ref{fig:synth_bpl}-A. These type of changes are compatible with human handwriting variability. 

From the out-of-sample examples in Fig.~\ref{fig:synth_bpl}-B, we can observe a much higher range of variations. The increase in variations is shown in both levels of type and token. At the token level, we can detect more diversity in affine transformations such as rotation (in third column of figure), scaling (third row of figure), translation in terms of the center. We also see a lot more diversity within symbols compared to the in-sample examples.

Apart from being realistic and introducing variety, what BPL offers cannot be obtained with other data generation techniques. Since BPL has used the actual human handwriting distribution, it is quite hard to reach a similar realism with any other ad-hoc generation technique.

\subsubsection{Human Evaluation}

Aside from providing qualitative results, we want to quantify the effectiveness of BPL in terms of how similar it is to its original examples. However, quantifying similarity in handwritten text is difficult. 
For this reason, we have run a human study following simplified version of our task formulation: Given an original symbol (query) and 5 options, human subjects have to choose the option that matches the query. In the experiment setup, 4 out of 5 options are BPL generated and a final option is ``Not Sure''.

\begin{figure}[t!]
    \centering
    \includegraphics[width=0.9\columnwidth,height=4.5cm]{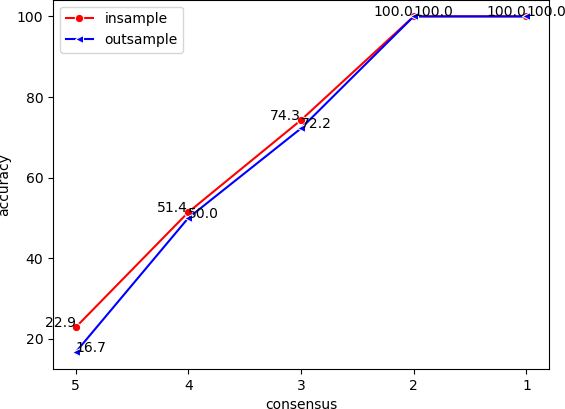}
    \caption{Result  of the AMT human study where subjects are asked to match between real and generated images. The consensus seen in the $x$-axis represents the amount of agreement among subjects.}
    \label{fig:human_study}
\end{figure}

We set up 2 experiments to quantify the ``realistic''generation: in-sample and out-of-sample generation. Both experiments follow the same procedure in which we provide Amazon Mechanical Turk (AMT) workers a single symbol (a query) and ask them to find the most visually similar item to the query.  
In the first experiment, we pick the options generated from the query while for the second experiment options can be generated from any symbol but the query. The former experimentation will provide how accurate BPL is within in-sample distribution and the latter is how well it can match to the out-of-sample distribution.
For each class, we have selected 5 original symbols and 2 BPL generated ones, giving us 10 task per symbol. Both experiments are set up using AMT, in which 5 workers had to answer each question.
In total, for each experiment we have 210 questions and 1050 answers from 5 different human subjects. 
The results of these experiments are shown in Fig.~\ref{fig:human_study}. We show the accuracy vs at least how many subjects chose the correct option. In other words, we plot what is the accuracy of at least $n$ workers choosing the correct option.   
As it can be deduced from Fig.~\ref{fig:human_study}, from at least 5 workers to at least 3 workers, there is a steady increase. Moreover, it is quite remarkable to observe that an ad-hoc method that requires no training can still result in 22.9\% or 16.7\% accuracy for all workers correctly predicting. 

However, we choose to focus on at least 3 workers correctly predicting because of the majority voting paradigm. Thus, according to at least 3 workers choosing correct option, we get 74.3\% and 72.2\% accuracy for in-sample and out-of-sample. The first conclusion is that we have quantified our method's accuracy and it is reasonably high given that the probability of randomly selecting correct option is 20\%. The second conclusion is that there is no much difference between in-sample and out-of-sample accuracy, only 2.1\%, which is encouraging considering the training procedure of our models.
Finally, we can see that if we relax the assumption of majority voting, we are getting 100\% accuracy for both experiments. This is quite promising since we have at least 2 human subjects that can match BPL samples to the query in all tasks in both types of experimentation. 

\section{Impact of Data Generation for HTR}\label{section:bpl_in_htr}
In this section we describe the generation of text lines, the HTR methods, and the evaluation on cipher text recognition.

\subsection{Data Generation results - Line Level}\label{section:linelevel}

Most HTR methods recognize the text at word or line level, because it is hard to segment it into isolated characters, which is also the case of most cipher manuscripts. As a consequence, segmenting symbols and classifying symbols is not a feasible option.
For this reason, we have created text lines to be used for training the HTR. Concretely, we take the symbols generated by BPL, and horizontally concatenate them in a manner as much realistic as possible. We set the space between characters, chosen randomly between $0$ and $30$ pixels, and we rotate each character randomly between $-5$ and $5$ degrees. Also, we add some artifacts to the upper part and lower part of the line. The synthetic text lines created from the BPL generated symbols are called the BPLL set. 

\begin{figure}[t]
    \small
    \begin{center}
    \begin{tabular}{p{0.8\linewidth}}
      \hline
      \multicolumn{1}{|l|}{\includegraphics[width=0.9\columnwidth, height=0.75cm]{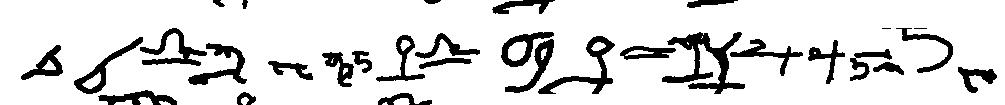}}\\
      \hline\noalign{\smallskip}
      Line created from the BPL generates symbols (BPLL set)\\
      \hline
      \multicolumn{1}{|l|}{\includegraphics[width=0.9\columnwidth, height=0.75cm]{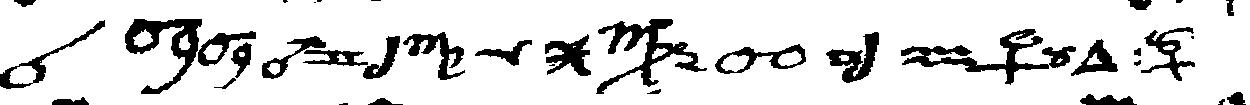}}\\
      \hline\noalign{\smallskip}
      Line created from the real few symbols + applied transformations (DAL set)\\
      \hline
      \multicolumn{1}{|l|}{\includegraphics[width=0.9\columnwidth, height=0.75cm]{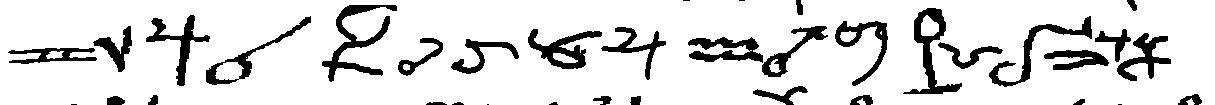}}\\
      \hline\noalign{\smallskip}
      Line created from mixing the above ones.
    \end{tabular}
    \end{center}
    \vspace{-0.5cm}
    \caption{Examples of the three sets of  lines, created by concatenating the symbols.}
    \label{fig:createdlines}
\end{figure}

For comparison, we also applied some data augmentation techniques 
(rotation, resizing, random thickness,...) on the real symbols and concatenated them to generated synthetic lines from the real symbols. We denote this set as DAL. Moreover, we created another set of lines by randomly mixing symbols from the two previous sets, resulting in three different sets of lines. Few samples from those lines are shown in Fig. \ref{fig:createdlines}. As it can be seen, some noise was introduced to make the lines as similar as possible to the real ones.  
The training of the HTR models can be done using one of the created sets, and also by mixing sets. Thus, we used the following two forms of mixing:

\begin{itemize}
    \item \textbf{Homogeneous Lines (HomL):} composed of lines from the BPLL set + lines created from the DAL set. In other words, we do not mix in the same line real symbols with BPL generated symbols and vice versa.
    
    \item \textbf{Heterogeneous Lines (HetL):}  composed of  lines created from the mixed symbols (generated by BPL and data augmentation) + lines from the DAL set. In other words, we mix real symbols with BPL generated symbols while generating a line and vice versa.
\end{itemize}

It is to note that we used three scenarios in this study: 10 samples, 5 samples and 1 sample. That means that we start with only 10, 5, or 1 example(s) of each of the Borg symbols, respectively, to perform the data augmentation and the BPL generation in order to create the synthetic data from a low resource alphabet.


\subsection{HTR models and Evaluation Metric}

After generating the synthetic lines, a HTR model can be trained for recognition. For this, we consider two options: A supervised model based on sequence to sequence with attention \cite{kang2018convolve} and a few-shot learning based model \cite{souibgui2020}. 

\textbf{Seq2Seq for HTR}
The first method follows most of the HTR models, where the goal is to learn a mapping function from  a  line image $X$ to  a text $Y$.  It is  an attention based sequence-to-sequence model,  proposed in \cite{kang2018convolve} and composed of three main parts: an encoder includes  a CNN and a bi-directional Gated Recurrent Unit (GRU), an attention mechanism and a decoder constituted from a one-directional GRU. Thus, given a line image as an input, the recognition is done character by character using the attention mechanism to produce the output text. This model showed competitive results in handwritten recognition with a huge amount of data using various Latin manuscript datasets for evaluation.

\textbf{Few-shot for HTR}
Since we are using one or few examples of each Borg symbols to generate the data (a few-shot generation), using a few-shot model for the recognition could be suitable. Thus, we choose the approach proposed in \cite{souibgui2020}, a segmentation free (works at line level) method for historical ciphered handwritten text recognition. It consists in inputting a ciphered text line image with an associated alphabet as isolated symbols to a matching model, where one or few examples (usually up to five) of each symbol should be given. Then, a similarity matrix  between the line and the alphabet is outputted. After that, the recognized text is decoded from the matrix.  Formally, if the size of the Borg alphabet is $N$ and we provide $k$ examples from each of the alphabet symbol for matching (the shots), the process is considered a $N$-way $k$-shot detection problem.

\begin{table*}[h]
\small
\centering
\caption{Obtained results by different methods and settings: Real and synthetic data were tested with with various sizes (\# of ann. lines). \# of generated samples indicates the number of images per each symbol, used to generate the synthetic lines.}
\label{tab:all_experiments}   
\begin{tabular}{l@{\hskip 0.2in}l@{\hskip 0.2in}l@{\hskip 0.2in}l@{\hskip 0.2in}l@{\hskip 0.2in}l}
\hline\noalign{\smallskip}

\textbf{Data Type} & \textbf{Model} & \textbf{\#of ann. lines}&\textbf{\# of gen. samples}&\textbf{$k$-shot} & \textbf{SER} \\

\noalign{\smallskip}\hline\noalign{\smallskip}
\multirow{7}{*}{ Real}& Unsupervised \cite{baro2019} & None&--&-- & 0.54 \\
\noalign{\smallskip} \cline{2-6}\noalign{\smallskip}
&\multirow{1}{*}{Few-shot \cite{souibgui2020}} & None&--&5 & {0.53} \\
\noalign{\smallskip} \cline{2-6}\noalign{\smallskip}
&\multirow{4}{*}{ MDLSTM \cite{fornes2017} }& $\approx$ 81 &--&--&0.71\\
&&$\approx$ 114&-- &--&0.66\\
&&$\approx$ 148&-- &--&0.69\\
&&$\approx$ 214&-- &--&0.55\\
\noalign{\smallskip} \cline{2-6}\noalign{\smallskip}
&\multirow{1}{*}{Few-shot \cite{souibgui2020}} & 117&-- &5 & \textbf{0.21} \\

\noalign{\smallskip}\hline\hline\noalign{\smallskip}

\multirow{3}{*}{Ours (HomL)}&\multirow{3}{*}{Few-shot \cite{souibgui2020}} &1000&10& 5 & 0.25 \\
&&1000&5& 5 & \textbf{0.25} \\
&&1000 &1& 1 & 0.31 \\

\noalign{\smallskip}\hline\noalign{\smallskip}
\multirow{3}{*}{Ours (HetL)}&\multirow{3}{*}{Few-shot \cite{souibgui2020}} &1000 &10& 5 & 0.30 \\
&&1000 &5& 5 & 0.28 \\
&&1000 &1& 1 & 0.41 \\

\noalign{\smallskip}\hline\noalign{\smallskip}
\multirow{3}{*}{Ours (HomL)}&\multirow{3}{*}{ \vtop{\hbox{\strut Seq2Seq +}\hbox{\strut Attention  \cite{kang2018convolve}}} }&1000 &10& - & 0.70 \\
&& 1000&5& - & 0.69 \\
&&1000 &1& - & 0.77 \\
\noalign{\smallskip}\hline\hline\noalign{\smallskip}
{Ours (HomL) + Real  } & Few-shot \cite{souibgui2020} & 117 + 117 &5 &5 & \textbf{0.20}\\

\noalign{\smallskip}\hline\noalign{\smallskip}
\multirow{3}{*}{Ours (HomL)}&\multirow{3}{*}{ \vtop{\hbox{\strut Seq2Seq +}\hbox{\strut Attention  \cite{kang2018convolve}}} }&2500 &5& - & 0.50 \\
&& 5000&5& - & 0.48 \\
&&10000 &5& - & 0.47 \\
&&20000 &5& - & 0.47 \\
\noalign{\smallskip}
\hline
\end{tabular}
\end{table*}

We have chosen this model because it has been applied to ciphers in a few shot scenario. The method is trained on synthetic alphabets (e.g. Omniglot \cite{lake2015human} constructed lines) and tested on real ciphered data, requiring only the support set. However, the results show that this model can obtain better results when it is fine-tuned on some real data.


\textbf{Evaluation Metric} The evaluation of the ciphered text transcription is done according to the Symbol Error Rate (SER) metric. It is similar to the Character Error Rate (CER) for text recognition. Formally, $SER = \frac{S+D+I}{N}$, where $S$, $D$ and $I$ are the numbers of required substitutions, deletions and  insertions, respectively, while  $N$ is  representing the  ground-truth's line length in term of symbols. 

\subsection{HTR Results}


We begin by finding the best setting to mix the created datasets, then we compare the best performance with the state of the art methods for transcribing the Borg ciphered manuscript using real  data.  

\begin{figure}[t]
     \setlength{\tabcolsep}{0pt}   
    
    \begin{tabular}{lll}
      \includegraphics[width=0.33\columnwidth]{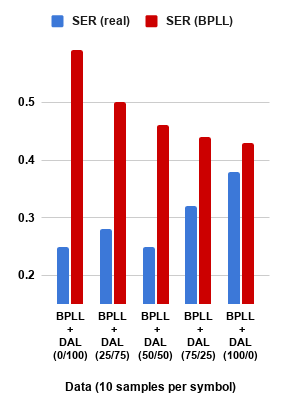}&
      \includegraphics[width=0.33\columnwidth]{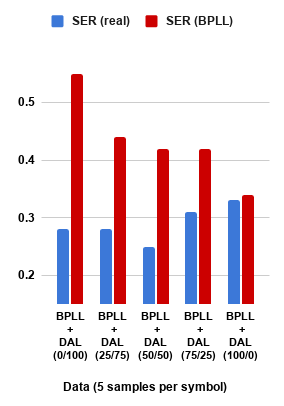}&
      \includegraphics[width=0.333\columnwidth]{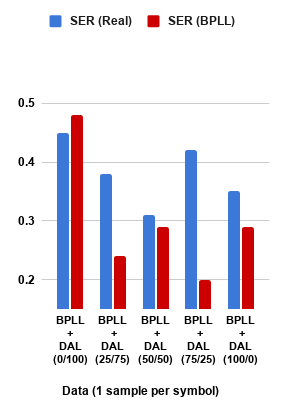}\\
    \end{tabular}
    \caption{SER of testing with  real Borg lines and synthetic BPLL lines,  using different mixing settings and conditioning on  different numbers of samples for generation.}
    \label{fig:mixing}
    \vspace{-.5cm}
\end{figure}

\textbf{Effect of mixing:} We took lines from   BPLL and DAL sets to find the best mixing setting for the HomL mixed set, defined above. The assumption is that using this mixing we can obtain better results than using separate sets.  Because, as it can be seen from Fig.~\ref{fig:createdlines}, BPLL lines are visually rich: the writing style variation of each symbol is realistic, while DAL lines are visually similar to the real Borg lines.
To find the right amount of lines from each set that should be added to the total mix, we perform an experiment of varying the  percentages of the BPLL and the DAL, and calculating the SER at each time. The used model for this training is the one  presented in \cite{souibgui2020} which shows a good results for Borg manuscript. As shown in Fig.~\ref{fig:mixing}, the performance of the model trained using only the BPLL is not optimal, but it is stable using different amount of shots $[0.33,0.38]$. On the other hand, by using only the DAL based on classic data augmentation techniques, the performance decreases if we reduce the number of examples per each Borg symbol that are used to create the data (from 0.25 using 10 samples  to 0.45 using 1 sample). Mixing both sets, leads to a better performance, especially if the mixed data is composed of 50 \% from each set. Thus, we can conclude that  adding  the same number of DAL to the BPLL lines acts like a regularization technique. 
That is why we will keep this setting in the next experiments to compare our generated data with using a real Borg data to train models.

\textbf{SOTA results:} The obtained results using different approaches and data settings are presented in Table \ref{tab:all_experiments}. We compare the performance when using real Borg lines (where we can apply unsupervised or supervised learning), versus using our synthetically generated lines.
As expected, annotated data free approaches, such as the unsupervised \cite{baro2019} or few-shot without fine-tuning \cite{souibgui2020} (although providing one or few examples from each symbol to be used as supports is still needed) obtains very poor results. The reasons are the high degree of similarity among the Borg symbols and the difficult symbol segmentation in the unsupervised approach, whereas in the few-shot method the problem is the difference of distribution between the Borg dataset and the Omniglot one that was used to train. It is better, hence, to train the segmentation free models with annotated data. This leads to two options: Using a real annotated data or using our synthetically created one.

In case of having a high amount of annotated lines, training a MDLSTM in a supervised way could lead to a good result. But, in our experiments, since the maximum number of available annotated lines is 214, and knowing that the Borg manuscript has different handwriting styles, the results are still moderate. Note that, of course, the performance improves when providing more annotated pages, which require more user effort. With the few-shot method, however, a much better result of 0.21 SER is obtained with few annotated data (117 lines for fine-tuning the model pretrained on Omniglot). But, even the annotation of these 117 lines at symbol level (i.e. providing the bounding box of each one) is a time consuming task. Nota that when performing the human annotation experiment, we found that those lines require approximately 4 hours to be labeled.   

To reduce this effort, the same model is fine-tuned  with our synthetically created lines.   Using the HomL set, 
the results  are slightly diminished, from 0.21 to 0.25 as SER. But, we believe that this difference of 0.04 is not worth because it implies annotating 117 lines. Instead, by using our BPL-based approach, the user just needs to provide 5 examples of each Borg symbol. Moreover, we can obtain a SER of 0.31 which is also competitive, when using only 1 example per symbol, to generate the lines. This proves the effectiveness of our synthetically generated data in replacing the real one, with a huge gain in annotation effort, and a minimal decrease in recognition performance. We also notice that using 10 or 5 examples to generate data gives the same results when testing instead of improving. This can be explained by increasing the out of the sample matching, which may require using more data to cover it. We note also that using the other set (HetL) for training, leads to a minor performance. 

For comparison sake, we tested the Seq2Seq \cite{kang2018convolve} model, with the same lines. But results were unsatisfactory because it needs much more data to be trained than MDLSTM. Hence, we generated thousands of text lines for training using the 5 shots setting. However, we can see that the performance stabilizes after using a certain amount of lines. The reason is that the generated samples variation is quite limited, since we are  only using 5 samples. Thus, we can conclude that generating fewer samples and training the few-shot model is a much better option.

Finally, we can see that when mixing the real Borg lines with the same amount of lines generated synthetically from 5 examples by BPL, we obtain the best result, concretely 0.20 SER, which indeed outperforms the state of the art of recognizing this manuscript.

\subsection{Latin Handwritten Text}

Next, we select a modern manuscript to further investigate the applicability of BPL generation. For this purpose, we take the English manuscript from the IAM dataset and simulate the low resource scenario by taking only 73 lines belonging to the writer 552. Then, we cropped 1 example from each English character (upper and lower cases) to generate data using our method. After that, we train the few shot model in a 1 shot setting (note that we are not using any labeled text lines, we are just using 1 labeled example from each isolated character). The obtained results are shown in Table~\ref{tab:latin}. As it can be seen, applying our model without the BPL generation leads to 0.35 as CER. While by adding the BPLL lines, we boost the performance to 0.31 as CER, which demonstrates the utility of our method. 
\begin{table}[h]
    \centering
    \begin{tabular}{lcc}
    \hline
        \textbf{Data Type} &\textbf{Model} & \textbf{CER} \\
    \hline        
        DAL & Few-shot~\cite{souibgui2020} & 0.35\\
        HomL & Few-shot~\cite{souibgui2020} & 0.31\\
        \hline
    \end{tabular}
    \caption{The results on IAM dataset, simulating the low resource handwritten recognition. The numbers are in terms of character error rate (lower is better).}
    \label{tab:latin}
    \vspace{-0.8cm}
\end{table}


\section{Conclusion}\label{section:conclusion}
In this paper we have used a one shot approach for compositional handwritten text generation and we have demonstrated its effectiveness for low resource text recognition, as an example of sequence recognition. Although we have taken historical ciphered manuscript recognition as a study case, it can be applied on any other alphabet or script. 

Our method uses BPL to generate synthetic symbols from few real examples. Afterwards, synthetic lines were created to train machine learning algorithms for HTR. From the experiments, we can say that the created data leads to competitive results  compared to using a real annotated dataset, with a significantly reducing manual annotation effort by a huge margin. Moreover, we have achieved the state of the art in Borg ciphered text recognition when combining it with real data for training.

As a future work, we will investigate more realistic approaches to create text lines from the generated symbols, for instance the impaired domain translations methods. Moreover, we will investigate using the compositional generation to directly create words or lines instead of isolated symbols, with the possibility of applying this to different manuscripts. 

\section*{Acknowledgment}

This work has been partially supported by the Swedish Research Council
(grant 2018-06074, DECRYPT), the Spanish project RTI2018-095645-B-C21, the Ramon y Cajal Fellowship RYC-2014-16831, the CERCA Program / Generalitat de Catalunya and a UAB PHD scholarship (B18P0073).

{\small
\bibliographystyle{ieee_fullname}
\bibliography{egpaper}
}

\onecolumn 
\begin{center}
\textbf{\LARGE Supplementary Material}
\end{center}
\appendix


\section{BPL Characters Generation Results}
We show in what follows some characters generations by the BPL, these characters are belonging to different  ciphered and regular manuscripts. 
\subsection{Borg Ciphered Manuscript}

\begin{figure}[h]
\begin{center}

\begin{tabular}{lll}

\begin{tabular}{lll}
                      
\multicolumn{3}{c}{\includegraphics[width=0.08\columnwidth]{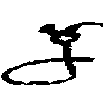}}                                                   \\ \hline
\multicolumn{1}{|l|}{\includegraphics[width=0.08\columnwidth]{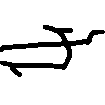}} & \multicolumn{1}{l|}{\includegraphics[width=0.08\columnwidth]{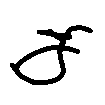}} & \multicolumn{1}{l|}{\includegraphics[width=0.08\columnwidth]{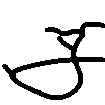}} \\ \hline
\multicolumn{1}{|l|}{\includegraphics[width=0.08\columnwidth]{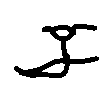}} & \multicolumn{1}{l|}{\includegraphics[width=0.08\columnwidth]{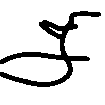}} & \multicolumn{1}{l|}{\includegraphics[width=0.08\columnwidth]{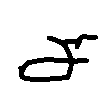}} \\ \hline
\multicolumn{1}{|l|}{\includegraphics[width=0.08\columnwidth]{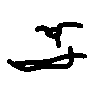}} & \multicolumn{1}{l|}{\includegraphics[width=0.08\columnwidth]{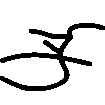}} & \multicolumn{1}{l|}{\includegraphics[width=0.08\columnwidth]{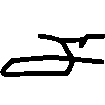}} \\ \hline
\end{tabular}
 &
\begin{tabular}{lll}
\multicolumn{3}{c}{\includegraphics[width=0.08\columnwidth]{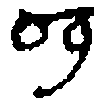}}                                                   \\ \hline
\multicolumn{1}{|l|}{\includegraphics[width=0.08\columnwidth]{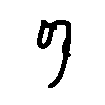}} & \multicolumn{1}{l|}{\includegraphics[width=0.08\columnwidth]{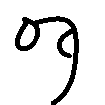}} & \multicolumn{1}{l|}{\includegraphics[width=0.08\columnwidth]{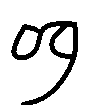}} \\ \hline
\multicolumn{1}{|l|}{\includegraphics[width=0.08\columnwidth]{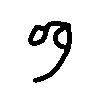}} & \multicolumn{1}{l|}{\includegraphics[width=0.08\columnwidth]{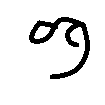}} & \multicolumn{1}{l|}{\includegraphics[width=0.08\columnwidth]{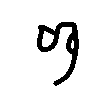}} \\ \hline
\multicolumn{1}{|l|}{\includegraphics[width=0.08\columnwidth]{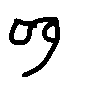}} & \multicolumn{1}{l|}{\includegraphics[width=0.08\columnwidth]{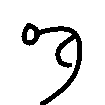}} & \multicolumn{1}{l|}{\includegraphics[width=0.08\columnwidth]{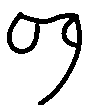}} \\ \hline
\end{tabular}

&

\begin{tabular}{lll}
\multicolumn{3}{c}{\includegraphics[width=0.08\columnwidth]{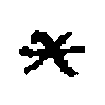}}                                                   \\ \hline
\multicolumn{1}{|l|}{\includegraphics[width=0.08\columnwidth]{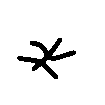}} & \multicolumn{1}{l|}{\includegraphics[width=0.08\columnwidth]{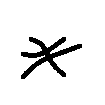}} & \multicolumn{1}{l|}{\includegraphics[width=0.08\columnwidth]{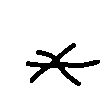}} \\ \hline
\multicolumn{1}{|l|}{\includegraphics[width=0.08\columnwidth]{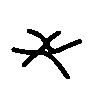}} & \multicolumn{1}{l|}{\includegraphics[width=0.08\columnwidth]{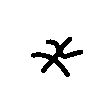}} & \multicolumn{1}{l|}{\includegraphics[width=0.08\columnwidth]{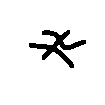}} \\ \hline
\multicolumn{1}{|l|}{\includegraphics[width=0.08\columnwidth]{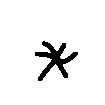}} & \multicolumn{1}{l|}{\includegraphics[width=0.08\columnwidth]{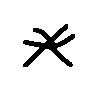}} & \multicolumn{1}{l|}{\includegraphics[width=0.08\columnwidth]{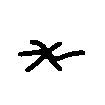}} \\ \hline
\end{tabular}

\cr
\noalign{\smallskip}\hline\noalign{\smallskip}

\begin{tabular}{lll}
                      \hline
\multicolumn{1}{|l|}{\includegraphics[width=0.08\columnwidth]{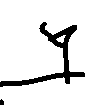}} & \multicolumn{1}{l|}{\includegraphics[width=0.08\columnwidth]{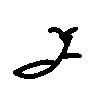}} & \multicolumn{1}{l|}{\includegraphics[width=0.08\columnwidth]{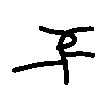}} \\ \hline
\multicolumn{1}{|l|}{\includegraphics[width=0.08\columnwidth]{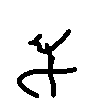}} & \multicolumn{1}{l|}{\includegraphics[width=0.08\columnwidth]{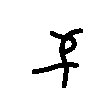}} & \multicolumn{1}{l|}{\includegraphics[width=0.08\columnwidth]{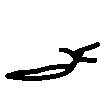}} \\ \hline
\multicolumn{1}{|l|}{\includegraphics[width=0.08\columnwidth]{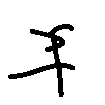}} & \multicolumn{1}{l|}{\includegraphics[width=0.08\columnwidth]{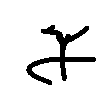}} & \multicolumn{1}{l|}{\includegraphics[width=0.08\columnwidth]{supp_images/borg_gen/q/outsample/7_20.png}} \\ \hline
\end{tabular}
 &
\begin{tabular}{lll}
                      \hline

\multicolumn{1}{|l|}{\includegraphics[width=0.08\columnwidth]{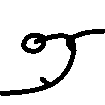}} & \multicolumn{1}{l|}{\includegraphics[width=0.08\columnwidth]{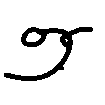}} & \multicolumn{1}{l|}{\includegraphics[width=0.08\columnwidth]{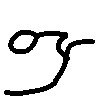}} \\ \hline
\multicolumn{1}{|l|}{\includegraphics[width=0.08\columnwidth]{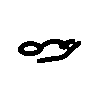}} & \multicolumn{1}{l|}{\includegraphics[width=0.08\columnwidth]{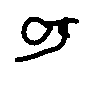}} & \multicolumn{1}{l|}{\includegraphics[width=0.08\columnwidth]{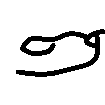}} \\ \hline
\multicolumn{1}{|l|}{\includegraphics[width=0.08\columnwidth]{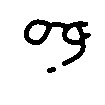}} & \multicolumn{1}{l|}{\includegraphics[width=0.08\columnwidth]{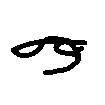}} & \multicolumn{1}{l|}{\includegraphics[width=0.08\columnwidth]{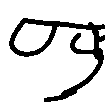}} \\ \hline
\end{tabular}

&

\begin{tabular}{lll}
                      \hline
\multicolumn{1}{|l|}{\includegraphics[width=0.08\columnwidth]{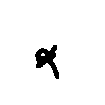}} & \multicolumn{1}{l|}{\includegraphics[width=0.08\columnwidth]{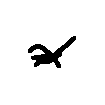}} & \multicolumn{1}{l|}{\includegraphics[width=0.08\columnwidth]{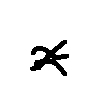}} \\ \hline
\multicolumn{1}{|l|}{\includegraphics[width=0.08\columnwidth]{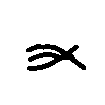}} & \multicolumn{1}{l|}{\includegraphics[width=0.08\columnwidth]{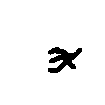}} & \multicolumn{1}{l|}{\includegraphics[width=0.08\columnwidth]{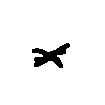}} \\ \hline
\multicolumn{1}{|l|}{\includegraphics[width=0.08\columnwidth]{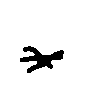}} & \multicolumn{1}{l|}{\includegraphics[width=0.08\columnwidth]{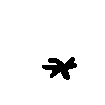}} & \multicolumn{1}{l|}{\includegraphics[width=0.08\columnwidth]{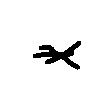}} \\ \hline
\end{tabular}

\end{tabular}
\end{center}
\caption{Generating new exemplars given one Borg ciphered symbol. \textbf{(UP):}  Conditioning on the same symbol (in-sample) shown on top of the nine-cipher grids. \textbf{(DOWN):} Conditioning on a different example of the same class (out-sample). The nine-character grids were generated by BPL.}
\label{fig:synth_bpl}
\end{figure}


\subsection{Chinese Manuscript}

\begin{figure}[H]
\begin{center}

\begin{tabular}{lll}

\begin{tabular}{lll}
                      
\multicolumn{3}{c}{\includegraphics[width=0.08\columnwidth]{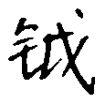}}                                                   \\ \hline
\multicolumn{1}{|l|}{\includegraphics[width=0.08\columnwidth]{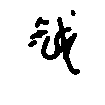}} & \multicolumn{1}{l|}{\includegraphics[width=0.08\columnwidth]{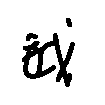}} & \multicolumn{1}{l|}{\includegraphics[width=0.08\columnwidth]{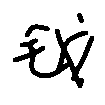}} \\ \hline
\multicolumn{1}{|l|}{\includegraphics[width=0.08\columnwidth]{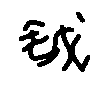}} & \multicolumn{1}{l|}{\includegraphics[width=0.08\columnwidth]{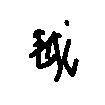}} & \multicolumn{1}{l|}{\includegraphics[width=0.08\columnwidth]{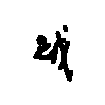}} \\ \hline
\multicolumn{1}{|l|}{\includegraphics[width=0.08\columnwidth]{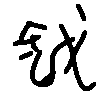}} & \multicolumn{1}{l|}{\includegraphics[width=0.08\columnwidth]{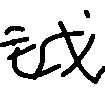}} & \multicolumn{1}{l|}{\includegraphics[width=0.08\columnwidth]{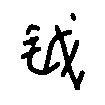}} \\ \hline
\end{tabular}
 &
\begin{tabular}{lll}
\multicolumn{3}{c}{\includegraphics[width=0.08\columnwidth]{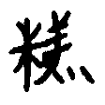}}                                                   \\ \hline
\multicolumn{1}{|l|}{\includegraphics[width=0.08\columnwidth]{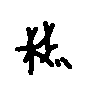}} & \multicolumn{1}{l|}{\includegraphics[width=0.08\columnwidth]{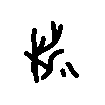}} & \multicolumn{1}{l|}{\includegraphics[width=0.08\columnwidth]{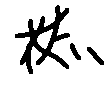}} \\ \hline
\multicolumn{1}{|l|}{\includegraphics[width=0.08\columnwidth]{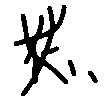}} & \multicolumn{1}{l|}{\includegraphics[width=0.08\columnwidth]{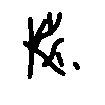}} & \multicolumn{1}{l|}{\includegraphics[width=0.08\columnwidth]{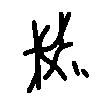}} \\ \hline
\multicolumn{1}{|l|}{\includegraphics[width=0.08\columnwidth]{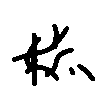}} & \multicolumn{1}{l|}{\includegraphics[width=0.08\columnwidth]{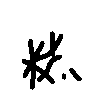}} & \multicolumn{1}{l|}{\includegraphics[width=0.08\columnwidth]{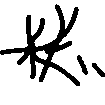}} \\ \hline
\end{tabular}

&

\begin{tabular}{lll}
\multicolumn{3}{c}{\includegraphics[width=0.08\columnwidth]{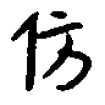}}                                                   \\ \hline
\multicolumn{1}{|l|}{\includegraphics[width=0.08\columnwidth]{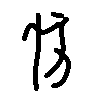}} & \multicolumn{1}{l|}{\includegraphics[width=0.08\columnwidth]{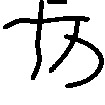}} & \multicolumn{1}{l|}{\includegraphics[width=0.08\columnwidth]{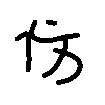}} \\ \hline
\multicolumn{1}{|l|}{\includegraphics[width=0.08\columnwidth]{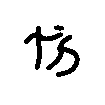}} & \multicolumn{1}{l|}{\includegraphics[width=0.08\columnwidth]{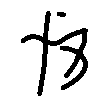}} & \multicolumn{1}{l|}{\includegraphics[width=0.08\columnwidth]{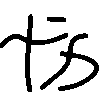}} \\ \hline
\multicolumn{1}{|l|}{\includegraphics[width=0.08\columnwidth]{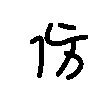}} & \multicolumn{1}{l|}{\includegraphics[width=0.08\columnwidth]{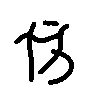}} & \multicolumn{1}{l|}{\includegraphics[width=0.08\columnwidth]{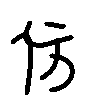}} \\ \hline
\end{tabular}

\cr
\noalign{\smallskip}\hline\noalign{\smallskip}

\begin{tabular}{lll}
                      \hline
\multicolumn{1}{|l|}{\includegraphics[width=0.08\columnwidth]{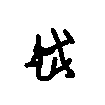}} & \multicolumn{1}{l|}{\includegraphics[width=0.08\columnwidth]{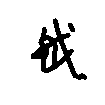}} & \multicolumn{1}{l|}{\includegraphics[width=0.08\columnwidth]{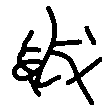}} \\ \hline
\multicolumn{1}{|l|}{\includegraphics[width=0.08\columnwidth]{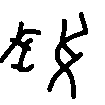}} & \multicolumn{1}{l|}{\includegraphics[width=0.08\columnwidth]{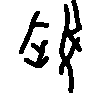}} & \multicolumn{1}{l|}{\includegraphics[width=0.08\columnwidth]{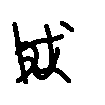}} \\ \hline
\multicolumn{1}{|l|}{\includegraphics[width=0.08\columnwidth]{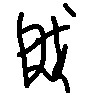}} & \multicolumn{1}{l|}{\includegraphics[width=0.08\columnwidth]{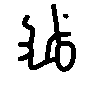}} & \multicolumn{1}{l|}{\includegraphics[width=0.08\columnwidth]{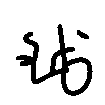}} \\ \hline
\end{tabular}
 &
\begin{tabular}{lll}
                      \hline

\multicolumn{1}{|l|}{\includegraphics[width=0.08\columnwidth]{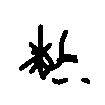}} & \multicolumn{1}{l|}{\includegraphics[width=0.08\columnwidth]{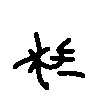}} & \multicolumn{1}{l|}{\includegraphics[width=0.08\columnwidth]{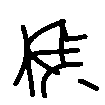}} \\ \hline
\multicolumn{1}{|l|}{\includegraphics[width=0.08\columnwidth]{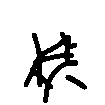}} & \multicolumn{1}{l|}{\includegraphics[width=0.08\columnwidth]{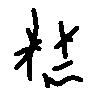}} & \multicolumn{1}{l|}{\includegraphics[width=0.08\columnwidth]{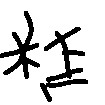}} \\ \hline
\multicolumn{1}{|l|}{\includegraphics[width=0.08\columnwidth]{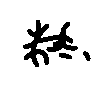}} & \multicolumn{1}{l|}{\includegraphics[width=0.08\columnwidth]{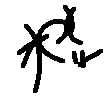}} & \multicolumn{1}{l|}{\includegraphics[width=0.08\columnwidth]{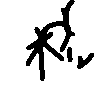}} \\ \hline
\end{tabular}

&

\begin{tabular}{lll}
                      \hline
\multicolumn{1}{|l|}{\includegraphics[width=0.08\columnwidth]{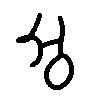}} & \multicolumn{1}{l|}{\includegraphics[width=0.08\columnwidth]{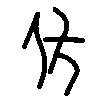}} & \multicolumn{1}{l|}{\includegraphics[width=0.08\columnwidth]{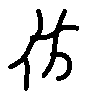}} \\ \hline
\multicolumn{1}{|l|}{\includegraphics[width=0.08\columnwidth]{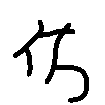}} & \multicolumn{1}{l|}{\includegraphics[width=0.08\columnwidth]{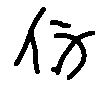}} & \multicolumn{1}{l|}{\includegraphics[width=0.08\columnwidth]{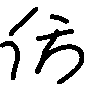}} \\ \hline
\multicolumn{1}{|l|}{\includegraphics[width=0.08\columnwidth]{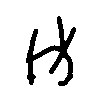}} & \multicolumn{1}{l|}{\includegraphics[width=0.08\columnwidth]{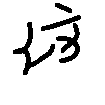}} & \multicolumn{1}{l|}{\includegraphics[width=0.08\columnwidth]{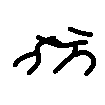}} \\ \hline
\end{tabular}

\end{tabular}
\end{center}
\caption{Generating new exemplars given one Chinese character. \textbf{(UP):}  Conditioning on the same character (in-sample) shown on top of the nine grids. \textbf{(DOWN):} Conditioning on a different example of the same class (out-sample). The nine-character grids were generated by BPL.}
\label{fig:synth_bpl}
\end{figure}

\subsection{Greek Manuscript}

\begin{figure}[H]
\begin{center}

\begin{tabular}{lll}

\begin{tabular}{lll}
                      
\multicolumn{3}{c}{\includegraphics[width=0.08\columnwidth]{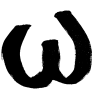}}                                                   \\ \hline
\multicolumn{1}{|l|}{\includegraphics[width=0.08\columnwidth]{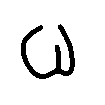}} & \multicolumn{1}{l|}{\includegraphics[width=0.08\columnwidth]{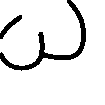}} & \multicolumn{1}{l|}{\includegraphics[width=0.08\columnwidth]{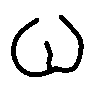}} \\ \hline
\multicolumn{1}{|l|}{\includegraphics[width=0.08\columnwidth]{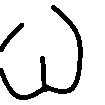}} & \multicolumn{1}{l|}{\includegraphics[width=0.08\columnwidth]{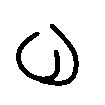}} & \multicolumn{1}{l|}{\includegraphics[width=0.08\columnwidth]{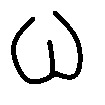}} \\ \hline
\multicolumn{1}{|l|}{\includegraphics[width=0.08\columnwidth]{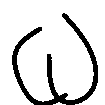}} & \multicolumn{1}{l|}{\includegraphics[width=0.08\columnwidth]{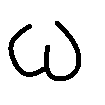}} & \multicolumn{1}{l|}{\includegraphics[width=0.08\columnwidth]{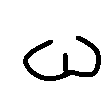}} \\ \hline
\end{tabular}
 &
\begin{tabular}{lll}
\multicolumn{3}{c}{\includegraphics[width=0.08\columnwidth]{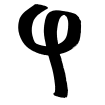}}                                                   \\ \hline
\multicolumn{1}{|l|}{\includegraphics[width=0.08\columnwidth]{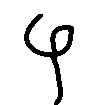}} & \multicolumn{1}{l|}{\includegraphics[width=0.08\columnwidth]{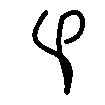}} & \multicolumn{1}{l|}{\includegraphics[width=0.08\columnwidth]{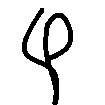}} \\ \hline
\multicolumn{1}{|l|}{\includegraphics[width=0.08\columnwidth]{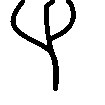}} & \multicolumn{1}{l|}{\includegraphics[width=0.08\columnwidth]{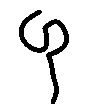}} & \multicolumn{1}{l|}{\includegraphics[width=0.08\columnwidth]{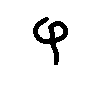}} \\ \hline
\multicolumn{1}{|l|}{\includegraphics[width=0.08\columnwidth]{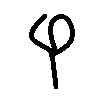}} & \multicolumn{1}{l|}{\includegraphics[width=0.08\columnwidth]{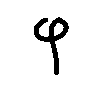}} & \multicolumn{1}{l|}{\includegraphics[width=0.08\columnwidth]{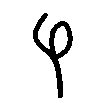}} \\ \hline
\end{tabular}

&

\begin{tabular}{lll}
\multicolumn{3}{c}{\includegraphics[width=0.08\columnwidth]{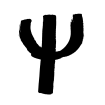}}                                                   \\ \hline
\multicolumn{1}{|l|}{\includegraphics[width=0.08\columnwidth]{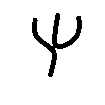}} & \multicolumn{1}{l|}{\includegraphics[width=0.08\columnwidth]{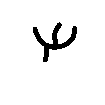}} & \multicolumn{1}{l|}{\includegraphics[width=0.08\columnwidth]{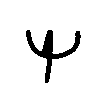}} \\ \hline
\multicolumn{1}{|l|}{\includegraphics[width=0.08\columnwidth]{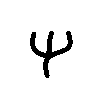}} & \multicolumn{1}{l|}{\includegraphics[width=0.08\columnwidth]{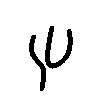}} & \multicolumn{1}{l|}{\includegraphics[width=0.08\columnwidth]{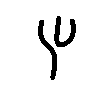}} \\ \hline
\multicolumn{1}{|l|}{\includegraphics[width=0.08\columnwidth]{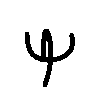}} & \multicolumn{1}{l|}{\includegraphics[width=0.08\columnwidth]{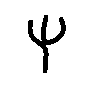}} & \multicolumn{1}{l|}{\includegraphics[width=0.08\columnwidth]{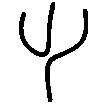}} \\ \hline
\end{tabular}

\cr
\noalign{\smallskip}\hline\noalign{\smallskip}

\begin{tabular}{lll}
                      \hline
\multicolumn{1}{|l|}{\includegraphics[width=0.08\columnwidth]{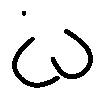}} & \multicolumn{1}{l|}{\includegraphics[width=0.08\columnwidth]{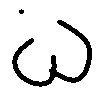}} & \multicolumn{1}{l|}{\includegraphics[width=0.08\columnwidth]{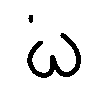}} \\ \hline
\multicolumn{1}{|l|}{\includegraphics[width=0.08\columnwidth]{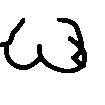}} & \multicolumn{1}{l|}{\includegraphics[width=0.08\columnwidth]{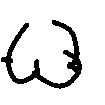}} & \multicolumn{1}{l|}{\includegraphics[width=0.08\columnwidth]{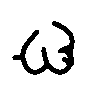}} \\ \hline
\multicolumn{1}{|l|}{\includegraphics[width=0.08\columnwidth]{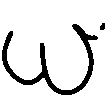}} & \multicolumn{1}{l|}{\includegraphics[width=0.08\columnwidth]{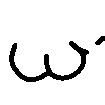}} & \multicolumn{1}{l|}{\includegraphics[width=0.08\columnwidth]{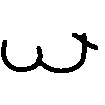}} \\ \hline
\end{tabular}
 &
\begin{tabular}{lll}
                      \hline

\multicolumn{1}{|l|}{\includegraphics[width=0.08\columnwidth]{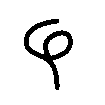}} & \multicolumn{1}{l|}{\includegraphics[width=0.08\columnwidth]{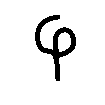}} & \multicolumn{1}{l|}{\includegraphics[width=0.08\columnwidth]{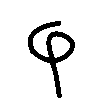}} \\ \hline
\multicolumn{1}{|l|}{\includegraphics[width=0.08\columnwidth]{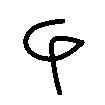}} & \multicolumn{1}{l|}{\includegraphics[width=0.08\columnwidth]{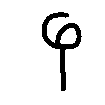}} & \multicolumn{1}{l|}{\includegraphics[width=0.08\columnwidth]{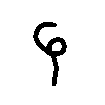}} \\ \hline
\multicolumn{1}{|l|}{\includegraphics[width=0.08\columnwidth]{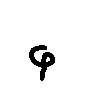}} & \multicolumn{1}{l|}{\includegraphics[width=0.08\columnwidth]{supp_images/gree_gen/4/outsample/26_20.png}} & \multicolumn{1}{l|}{\includegraphics[width=0.08\columnwidth]{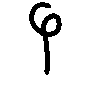}} \\ \hline
\end{tabular}

&

\begin{tabular}{lll}
                      \hline
\multicolumn{1}{|l|}{\includegraphics[width=0.08\columnwidth]{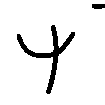}} & \multicolumn{1}{l|}{\includegraphics[width=0.08\columnwidth]{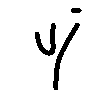}} & \multicolumn{1}{l|}{\includegraphics[width=0.08\columnwidth]{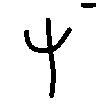}} \\ \hline
\multicolumn{1}{|l|}{\includegraphics[width=0.08\columnwidth]{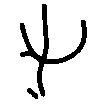}} & \multicolumn{1}{l|}{\includegraphics[width=0.08\columnwidth]{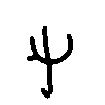}} & \multicolumn{1}{l|}{\includegraphics[width=0.08\columnwidth]{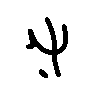}} \\ \hline
\multicolumn{1}{|l|}{\includegraphics[width=0.08\columnwidth]{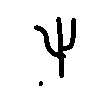}} & \multicolumn{1}{l|}{\includegraphics[width=0.08\columnwidth]{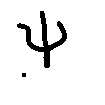}} & \multicolumn{1}{l|}{\includegraphics[width=0.08\columnwidth]{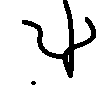}} \\ \hline
\end{tabular}

\end{tabular}
\end{center}
\caption{Generating new exemplars given one Greek character. \textbf{(UP):}  Conditioning on the same character (in-sample) shown on top of the nine grids. \textbf{(DOWN):} Conditioning on a different example of the same class (out-sample). The nine-character grids were generated by BPL.}
\label{fig:synth_bpl}
\end{figure}

\subsection{Latin Manuscript}

\begin{figure}[H]
\begin{center}

\begin{tabular}{lll}

\begin{tabular}{lll}
                      
\multicolumn{3}{c}{\includegraphics[width=0.08\columnwidth]{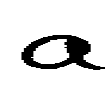}}                                                   \\ \hline
\multicolumn{1}{|l|}{\includegraphics[width=0.08\columnwidth]{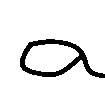}} & \multicolumn{1}{l|}{\includegraphics[width=0.08\columnwidth]{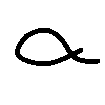}} & \multicolumn{1}{l|}{\includegraphics[width=0.08\columnwidth]{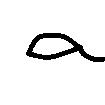}} \\ \hline
\multicolumn{1}{|l|}{\includegraphics[width=0.08\columnwidth]{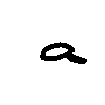}} & \multicolumn{1}{l|}{\includegraphics[width=0.08\columnwidth]{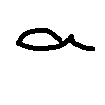}} & \multicolumn{1}{l|}{\includegraphics[width=0.08\columnwidth]{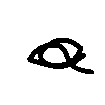}} \\ \hline
\multicolumn{1}{|l|}{\includegraphics[width=0.08\columnwidth]{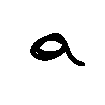}} & \multicolumn{1}{l|}{\includegraphics[width=0.08\columnwidth]{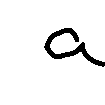}} & \multicolumn{1}{l|}{\includegraphics[width=0.08\columnwidth]{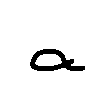}} \\ \hline
\end{tabular}
 &
\begin{tabular}{lll}
\multicolumn{3}{c}{\includegraphics[width=0.08\columnwidth]{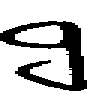}}                                                   \\ \hline
\multicolumn{1}{|l|}{\includegraphics[width=0.08\columnwidth]{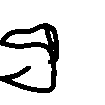}} & \multicolumn{1}{l|}{\includegraphics[width=0.08\columnwidth]{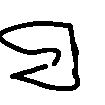}} & \multicolumn{1}{l|}{\includegraphics[width=0.08\columnwidth]{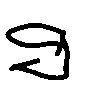}} \\ \hline
\multicolumn{1}{|l|}{\includegraphics[width=0.08\columnwidth]{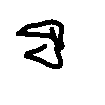}} & \multicolumn{1}{l|}{\includegraphics[width=0.08\columnwidth]{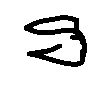}} & \multicolumn{1}{l|}{\includegraphics[width=0.08\columnwidth]{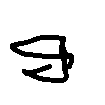}} \\ \hline
\multicolumn{1}{|l|}{\includegraphics[width=0.08\columnwidth]{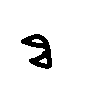}} & \multicolumn{1}{l|}{\includegraphics[width=0.08\columnwidth]{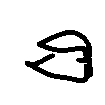}} & \multicolumn{1}{l|}{\includegraphics[width=0.08\columnwidth]{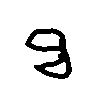}} \\ \hline
\end{tabular}

&

\begin{tabular}{lll}
\multicolumn{3}{c}{\includegraphics[width=0.08\columnwidth]{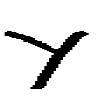}}                                                   \\ \hline
\multicolumn{1}{|l|}{\includegraphics[width=0.08\columnwidth]{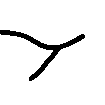}} & \multicolumn{1}{l|}{\includegraphics[width=0.08\columnwidth]{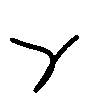}} & \multicolumn{1}{l|}{\includegraphics[width=0.08\columnwidth]{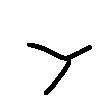}} \\ \hline
\multicolumn{1}{|l|}{\includegraphics[width=0.08\columnwidth]{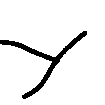}} & \multicolumn{1}{l|}{\includegraphics[width=0.08\columnwidth]{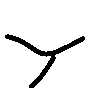}} & \multicolumn{1}{l|}{\includegraphics[width=0.08\columnwidth]{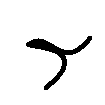}} \\ \hline
\multicolumn{1}{|l|}{\includegraphics[width=0.08\columnwidth]{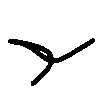}} & \multicolumn{1}{l|}{\includegraphics[width=0.08\columnwidth]{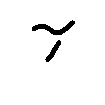}} & \multicolumn{1}{l|}{\includegraphics[width=0.08\columnwidth]{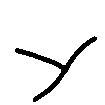}} \\ \hline
\end{tabular}

\end{tabular}
\end{center}
\caption{Generating new exemplars given one Latin character. The nine-character grids were generated by BPL.}
\label{fig:synth_bpl}
\end{figure}

\section{GAN Generations Results}
\begin{figure}[h]
\begin{center}


\begin{tabular}{lll}
                      
\multicolumn{3}{c}{\includegraphics[width=0.08\columnwidth]{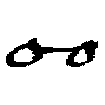}}                                                   \\ \hline
\multicolumn{1}{|l|}{\includegraphics[width=0.08\columnwidth]{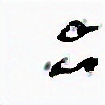}} & \multicolumn{1}{l|}{\includegraphics[width=0.08\columnwidth]{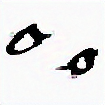}} & \multicolumn{1}{l|}{\includegraphics[width=0.08\columnwidth]{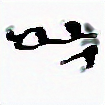}} \\ \hline
\multicolumn{1}{|l|}{\includegraphics[width=0.08\columnwidth]{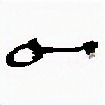}} & \multicolumn{1}{l|}{\includegraphics[width=0.08\columnwidth]{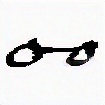}} & \multicolumn{1}{l|}{\includegraphics[width=0.08\columnwidth]{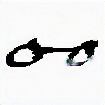}} \\ \hline
\multicolumn{1}{|l|}{\includegraphics[width=0.08\columnwidth]{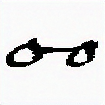}} & \multicolumn{1}{l|}{\includegraphics[width=0.08\columnwidth]{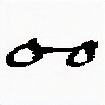}} & \multicolumn{1}{l|}{\includegraphics[width=0.08\columnwidth]{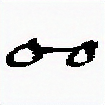}} \\ \hline
\end{tabular}
\end{center}
\caption{Generating new exemplars given one cipeherd symbol. The nine-character grids were generated by SinGan.}
\label{fig:singan}
\end{figure}

For comparison, We tested generating characters in a single shot scenario by GANs. For this purpose,  we used the SinGan \cite{shaham2019singan} that was developed to generate from a single image.  The obtained results are shown in Figure~\ref{fig:singan}, as it can be seen these are  unsatisfactory. The model starts by producing noisy images and then overfitted and produce the same image that was conditioned on.

\end{document}